%% file: main.tex
\def\year{2022}\relax
\documentclass[letterpaper]{article} % DO NOT CHANGE THIS
\usepackage{aaai22}  % DO NOT CHANGE THIS
\usepackage{times}  % DO NOT CHANGE THIS
\usepackage{helvet}  % DO NOT CHANGE THIS
\usepackage{courier}  % DO NOT CHANGE THIS
\usepackage[hyphens]{url}  % DO NOT CHANGE THIS
\usepackage{graphicx} % DO NOT CHANGE THIS
\usepackage{natbib}  % DO NOT CHANGE THIS AND DO NOT ADD ANY OPTIONS TO IT
\usepackage{caption} % DO NOT CHANGE THIS AND DO NOT ADD ANY OPTIONS TO IT
\DeclareCaptionStyle{ruled}{labelfont=normalfont,labelsep=colon,strut=off} % DO NOT CHANGE THIS
\usepackage[ruled,vlined]{algorithm2e}
\SetKwInput{KwParam}{Parameters}
\usepackage{tikz}
\usepackage{amsmath,amsfonts}
\usepackage[switch]{lineno}  

\newcommand{\eps}{\varepsilon}

\newcommand{\E}{\mathbb{E}}
\newcommand{\N}{\mathbb{N}}

\newcommand{\R}{\mathbb{R}}

\newcommand{\Ccal}{\mathcal{C}}

\newcommand{\Hcal}{\mathcal{H}}
\newcommand{\Kcal}{\mathcal{K}}

\newcommand{\Mcal}{\mathcal{M}}

\newcommand{\Zcal}{\mathcal{Z}}
\newcommand{\Wcal}{\mathcal{W}}

\newcommand{\vM}{\mathbf{M}}
\newcommand{\vh}{\mathbf{h}}

\usepackage{newfloat}
\usepackage{listings}
\title{SPATE-GAN: Improved Generative Modeling of Dynamic Spatio-Temporal Patterns with an Autoregressive Embedding Loss}
\author {
    Konstantin Klemmer,\equalcontrib \textsuperscript{\rm 1,2}
    Tianlin Xu, \equalcontrib \textsuperscript{\rm 3}
    Beatrice Acciaio, \textsuperscript{\rm 4}
    Daniel B Neill \textsuperscript{\rm 2}
}
\affiliations {
    \textsuperscript{\rm 1} University of Warwick \\
    \textsuperscript{\rm 2} New York University \\
    \textsuperscript{\rm 3} London School of Economics \\
    \textsuperscript{\rm 4} ETH Zurich \\
    \{k.klemmer, daniel.neill\}@nyu.edu, t.xu12@lse.ac.uk, beatrice.acciaio@math.ethz.ch
}
\usepackage{bibentry}
\begin{document}
%\linenumbers

\nocopyright
\maketitle

\begin{abstract}
From ecology to atmospheric sciences, many academic disciplines deal with data characterized by intricate spatio-temporal complexities, the modeling of which often requires specialized approaches. Generative models of these data are of particular interest, as they enable a range of impactful downstream applications like simulation or creating synthetic training data. Recent work has highlighted the potential of generative adversarial nets (GANs) for generating spatio-temporal data.  A new GAN algorithm COT-GAN, inspired by the theory of causal optimal transport (COT), was proposed in an attempt to better tackle this challenge.  However, the task of learning more complex spatio-temporal patterns requires additional knowledge of their specific data structures. 
In this study, we propose a novel loss objective combined with COT-GAN based on an autoregressive embedding to reinforce the learning of spatio-temporal dynamics. We devise SPATE (\textbf{spa}tio-\textbf{te}mporal association), a new metric measuring spatio-temporal autocorrelation by using the deviance of observations from their expected values. We compute SPATE for real and synthetic data samples and use it to compute an embedding loss that considers space-time interactions, nudging the GAN to learn outputs that are faithful to the observed dynamics. We test this new objective on a diverse set of complex spatio-temporal patterns: turbulent flows, log-Gaussian Cox processes and global weather data. We show that our novel embedding loss improves performance without any changes to the architecture of the COT-GAN backbone, highlighting our model's increased capacity for capturing autoregressive structures. We also contextualize our work with respect to recent advances in physics-informed deep learning and interdisciplinary work connecting neural networks with geographic and geophysical sciences.
%{\cred Bea's comments: both in the abstract and in the introduction we should at least mention that we use some newly developed modification of OT (or of the Earth Mover distance, if this terminology is more familiar to the targeted audience), suitable for sequential learning}

\end{abstract}

\section{Introduction}
\input{intro}

\section{Related work}
\input{related_work}

\section{Method}
\input{method}

\section{Experiments}\label{sec:experiments}
\input{experiments}

\section{Conclusion}
\input{conclusion}

%Use \bibliography{yourbibfile} instead or the References section will not appear in your paper
\bibliography{aaai22_konstantin,lin.bib}

\section{Acknowledgments}
The authors gratefully acknowledge funding from the UK Engineering and Physical Sciences Research Council, the EPSRC Centre for Doctoral Training in Urban Science (EPSRC grant no. EP/L016400/1).

\section{Appendix}
\input{appendix}

\end{document}

%% file: intro.tex
Over the last decade, deep learning has emerged as a powerful paradigm for modeling complex data structures. It has also found successful applications in the video domain, for example for trajectory forecasting, video super-resolution or object tracking. Nevertheless, data observed over (discrete) space and time can take many more shapes than just RGB videos: many of the systems and processes governing our planet, from ocean streams to the spread of viruses, exhibit complex spatio-temporal dynamics. Current deep learning approaches often struggle to account for these, as a recent survey by \citet{Reichstein2019} highlights. The authors call for more concerted research efforts aiming to improve the capacity of deep neural networks for modeling earth systems data. Recently, the emergence of physics-informed deep learning has reinforced the integration of physical constraints as a research domain \cite{Zhang2021,Rao2020,Wang2020,Kim2020a}. 

In this work, we propose a novel GAN tailored to the challenges of spatio-temporal complexities. We first devise a novel measure of spatio-temporal association---SPATE---expanding on the Moran's I measure of spatial autocorrelation. SPATE uses the deviance of an observation from its space-time expectation, and compares it to neighboring observations to identify regions of (relative) change and regions of (relative) homogeneity over time. We propose three different approaches to calculate the space-time expectations, coming with varying assumptions and advantages for different applications. %We then encode a SPATE-based embedding into an existing GAN framework, named SPATE-GAN, to reinforce the learning of spatio-temporal autocorrelation. 
We then encode a SPATE-based embedding into COT-GAN \cite{Xu2020} to formulate a new GAN framework, named SPATE-GAN.  The motivation of choosing COT-GAN as the base model is that its principle of respecting temporal dependencies in sequential modeling is in line with our intuition for SPATE,  see details in the methods section.
%We then construct a SPATE-based embedding loss to reinforce the learning of spatio-temporal autocorrelation into the model. Lastly, we test our approach on a range of different datasets. Specifically, we select data characterised by complex spatio-temporal patterns such as fluid dynamics \cite{Wallace2014,Wang2020}, disease spread \cite{Brix2001} or global surface temperatures \cite{Sismanidis2018}. We observe that our novel approach---SPATE-GAN---clearly outperforms existing methods. This finding is particularly interesting as we don't change the architecture of the GAN backbones, implying that our performance gains can be solely attributed to our novel embedding loss.
Lastly, we test our approach on a range of different datasets. Specifically, we select data characterized by complex spatio-temporal patterns such as fluid dynamics \cite{Wallace2014,Wang2020}, disease spread \cite{Brix2001} or global surface temperatures \cite{Sismanidis2018}. We observe that SPATE-GAN outperforms baseline models. This finding is particularly interesting as we do not change the architecture of the existing COT-GAN backbone, implying that our performance gains can be solely attributed to our novel SPATE-based embedding loss. 

To summarize, the contributions of this study are as follows:
\begin{itemize}
    \item We introduce SPATE, a new measure of spatio-temporal association, by expanding the intuition of the Moran's I metric into the temporal dimension.
    \item We introduce SPATE-GAN, a novel GAN for complex spatio-temporal data utilizing SPATE to construct an embedding loss "nudging" the model to focus on the learning of autoregressive structures.
    \item We test SPATE-GAN against baseline GANs designed for image/video generation on datasets representing fluid dynamics, disease spread and global surface temperature. We show performance gains of SPATE-GAN over the baseline models.
    %We test SPATE-GAN against baseline video GANs for generative modeling tasks on datasets representing fluid dynamics, disease spread and global surface temperature. We show substantial performance gains of SPATE-GAN over the baseline models.
\end{itemize}

%% file: related_work.tex
\subsection{Autocorrelation metrics for spatio-temporal phenomena}

Analysing autoregressive patterns in spatial and spatio-temporal data has a long tradition in different academic domains (e.g. GIS, ecology) which over time developed diverse measures to describe these phenomena. The most commonly known of these metrics is the Moran's I index of global and local spatial autocorrelation. Originally proposed by \citet{Anselin1995}, Moran's I identifies both homogeneous spatial clusters and outliers. Applications of the metric range from identifying rare earth contamination \cite{Yuan2018a} to analysing land cover change patterns \cite{Das2017}. Throughout the years, Moran's I has also motivated several methodological expansions, analysing for example spatial heteroskedasticity \cite{Ord2012} and local spatial dispersion \cite{Westerholt2018}. Moran's I has also seen some expansions into the spatio-temporal domain. 
\citet{Matthews2019} use the metric iteratively to model disease spread over time. \citet{Lee2017} and \citet{Gao2019} propose novel spatio-temporal expansions of the Moran's I metric, returning static outputs at a purely spatial resolution. \citet{Siino2018} design an extended Moran's I for spatio-temporal point processes. However, to the best of our knowledge, neither the Moran's I nor its spatio-temporal extensions have been applied to discrete spatio-temporal video data. It is evident that metrics of spatio-temporal autocorrelation can provide meaningful embeddings of complex data, capturing underlying patterns throughout a range of different application domains. 

\subsection{Deep learning \& GANs for spatial and spatio-temporal data}

Deep learning describes a powerful family of methods capable of dealing with the highly complex and non-linear nature of many real world spatial and spatio-temporal patterns \cite{Yan2017,Aodha2019,Chu2019,Bao2020,Geng2019}.  Paradigms like physics-informed deep learning aim to devise methods which integrate (geo)physical constraints explicitly into neural network models \cite{Wang2020}. There is also an increasing number of studies tackling specific challenges associated with geographic data: \citet{Mai2020} and \citet{Yin2019} propose context-aware vector embeddings for geographic coordinates. \citet{Zammit-Mangion2019} propose to learn deep neural networks for spatial covariance patterns using injective warping functions. Intuitions for spatial autocorrelation, including the Moran's I metric, have been integrated into machine learning frameworks tackling model selection for ensemble learners \cite{Klemmer2019a}, spatial representation learning \cite{Zhang2019}, auxiliary task learning \cite{Klemmer2021a} or residual correlation graph neural networks \cite{Jia2020}. All these studies highlight the benefits of explicitly encoding spatial context into neural networks to improve performance.

Narrowing down on the GAN context specifically, we find that spatio-temporal applications have mostly focused on video data \cite{Xu2020,Tulyakov2018,Kim2020}. Beyond this, GANs have been used for conditional density estimation of traffic \cite{Zhang2020a}, trajectory prediction \cite{Huang2021} or extreme weather event simulation \cite{Klemmer2021b}. Nevertheless, to the best of our knowledge, metrics capturing spatio-temporal autocorrelation have never been integrated into GANs. As previous studies highlight the value on encoding spatial context, this work seeks to provide a first-principle approach of integrating metrics of spatio-temporal autocorrelation into GANs for modeling of complex spatio-temporal patterns.

\subsection{Embedding loss functions}
As SPATE-GAN integrates spatio-temporal metrics into COT-GAN as embeddings, here we briefly revisit previous work and insights into embedding losses. Embedding losses describe approaches where the loss function of a model is computed on an embedding of the data. This can have desirable outcomes, such as training stability or focusing on specific patterns in the data. Embedding losses have become popular in computer vision over the last years: \citet{Ghafoorian2019} use embedding losses to improve GAN-based lane detection. \citet{Filntisis2020} use visual-semantic embedding losses to improve predictions of bodily expressed emotions. \citet{Wang2020a} introduce CLIFFNet, utilizing hierarchical embeddings of depth maps for molecular depth estimation. \citet{Bailer2017} introduce a threshold loss to improve optical flow estimation. It is clear that embedding losses have shown great potential for particularly challenging visual problems, especially those involving complex spatio-temporal dynamics. %We believe that the embedding loss is a natural choice for modelling spatio-temporal dynamics due to their complexities.

%% file: method.tex
\subsection{SPATE: Spatio-temporal association}

\begin{figure*}[ht]%{0.4\textwidth}
%\vskip -0.1in
\centering
\includegraphics[scale=0.99]{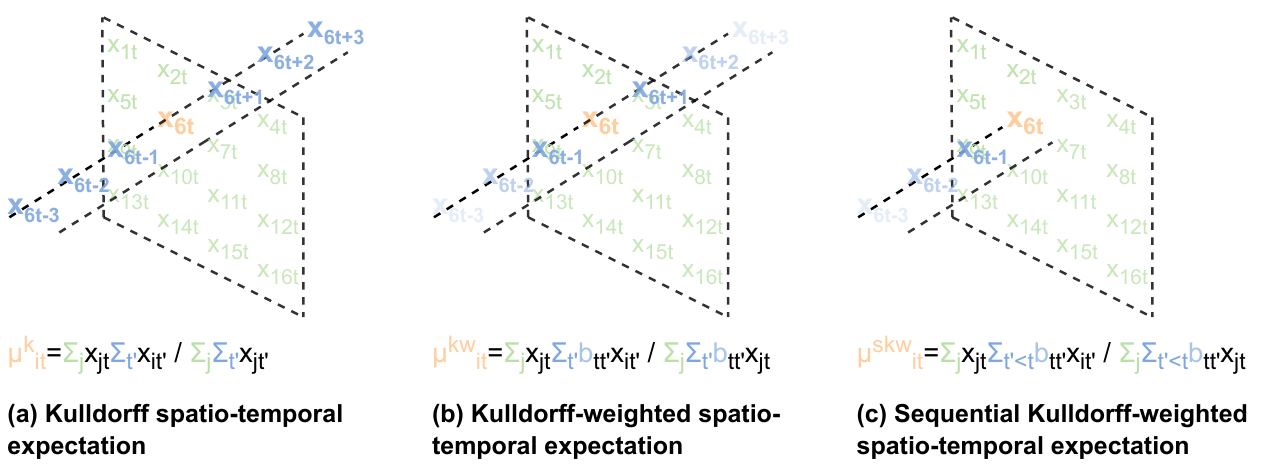}
%\includegraphics{fig1.pdf}
%\vskip -0.1in
\caption{Illustrating the three proposed options to obtain \textit{spatio-temporal expectations} $\mu_{it}$ used in the computation of $\text{SPATE}$ for single-channel data.}%, for (single-channel) video data.}
\label{fig1}
%\vskip -0.15in
%\vskip -0.5in
\end{figure*}

%{\cmgt L: what is a discrete spatial pattern with continuous values? Also, let's preserve "observation" for a data point or sequence $x$, and use "pixel" to refer to $x_i$.} {\cblue "Discrete" refers to a grid or other discrete spatial structure, as opposed to free-floating coordinates in continuous space. Sounds good RE the notation!}
For a static, discrete spatial pattern (e.g. a grid of pixels forming an image) consisting of continuous values $x \in \R^{n}$ where $n \in \mathbb{N}$ is the dimensionality of $x$, $x_i$ for $i \in 1,\dots,n$ represents the $i$'th pixel value on a regular grid.
% $x_i$ with $i \in 1,\dots,n$ (in our case continuous pixel values on a regular grid), 
The local Moran's I statistic $I_i$ is computed as follows:
\begin{equation}
\begin{split}
    I_{i} = (n_{i}-1) \frac{z_{i}}{\sum^{n_{i}}_{j=1} z_{j}^2} \sum^{n_{i}}_{j=1, j \neq i} w_{i,j} z_{j}
\end{split}
\end{equation}
where $z_i = x_{i} - \bar{x}$ is the deviance of observation $x_{i}$ from the global mean $\bar{x}$, %$n_x$ is the number of spatial neighbors of observation $i$ 
$n_{i}$ is the number of spatial neighbors of pixel $x_i$, $j$ indexes neighbors of $x_i$ for $j \in \{1, ..., n_i\}$ and $j \neq i$, and $w_{i,j}$ is a binary spatial weight matrix, indicating spatial neighborhood of observations $i$ and $j$. $I_{i}$ can be interpreted as a measure of similarity to neighboring %observations: positive values imply homogeneous clusters structure, while negative value suggest outliers, change patterns or edges.
pixels: positive values imply homogeneous clusters, while negative values suggest outliers, change patterns or edges.

Now, let us assume that we observe a sequence of spatial patterns over time $t$: $x=(x_1,...,x_T) \in \R^{n\times T}$ where $n$ is the dimensionality of $x_t$ at each time $t$ and $T$ is the total time steps of the sequence.
%Now, let us assume that we observe data not only in spatial units $i$ but also over time $t$. 
Of course, a naive adoption of the approach above is simply to ignore the time component of a sequence and compute the local Moran's I values  $I_{it}$ around pixel $i$ using mean values $\bar{x}_{t}$ at each time $t$. %using mean values $\bar{x}_{t}$. 
% We can still simply compute local Moran's I values  $I_{it}$ around the $i$'s pixel at each time $t$} using mean values $\bar{x}_{t}$. 
%{\cred (L: don't really understand the following sentence) These means can be thought of as \textit{spatial expectations}, that is assuming spatial independence (i.e. no spatial patterns), the expected value at a given spatial unit is simply the average of all spatial units.} {\cblue Yea we don't really need that sentence with the following one. Does it read ok like this?} 
Unfortunately, this approach would strictly separate spatial and temporal effects. In fact, the much more realistic assumption is that space and time are not separable, but do in fact interact and form joint patterns. For this reason, we expand the concept of Moran's I for spatio-temporal expectations. %in comparison to one that assumes space-time independence. 
First, we follow the intuition outlined in \citet{Kulldorff2005} and define expected values of $\mu_{it}(x)$.
% as the product of the sum of all values over time $t$ at unit $i$ and the sum of all spatial units $i$ at time step $t$, divided by the sum of all sum of all observations over all time steps. 
We refer to this approach as Kulldorff spatio-temporal expectation ("\textit{k}"):

\begin{equation} \label{eq:u_k}
\mu_{it}^{(k)} = \frac{\sum_{j} x_{jt} \sum_{t'} x_{it'}}{\sum_{j} \sum_{t'} x_{jt'}}.
\end{equation}

$\mu_{it}^{(k)}$ in \eqref{eq:u_k} involves utilizing all spatial units (pixels) available at time step $t$ and across all time steps at a single spatial unit (pixel position) $i$. This computation of the \textit{spatio-temporal expectations} 
assumes independence of space and time, and thus the residual
$z_{it} = x_{it} - \mu_{it}^{(k)}$ can be thought of as a local measure of space-time interaction at pixel $i$ and time $t$.  Moreover, this formulation of $\mu_{it}$ makes two critical assumptions: (1) Different time steps are equally important, irrespective of how distant they are from the current time step. (2) At each time step, we assume availability of the whole time series (i.e. looking into the future is possible). We can modify the computation of $\mu_{it}$ by imposing alternative assumptions.
%We can relax both assumptions by tweaking the computation of $\mu_{it}$ slightly. 

First, assuming that distant time steps have less significant impact on the current time step, we can integrate temporal weights into the computation, and apply decreasingly lower weights to more distant time steps. For example, one can consider an exponential kernel:

\begin{equation}
\mu_{it}^{(kw)} = \frac{\sum_j x_{jt} \sum_{t'} b_{tt'} x_{it'}}{\sum_j \sum_{t'} b_{tt'} x_{jt'}},
\end{equation}
where $b_{tt'} = \exp(-|t-t'| / l)$ and $l$ is the lengthscale of the exponential kernel. We refer to this approach as Kulldorff-weighted spatio-temporal expectation ("\textit{kw}").

Second, we can restrict the computation of $\mu_{it}$ at time step $t$ to only account for time steps $< t$, so that the expectation at each time step is independent of future observations. %, thus maintaining sequential logic:
Thus, the computation respects the generating logic of sequential data. We refer to this last approach as Kulldorff-sequential-weighted spatio-temporal expectation ("\textit{ksw}"):

\begin{equation}
\mu_{it}^{(ksw)} = \frac{\sum_j x_{jt} \sum_{t' < t} b_{tt'} x_{it'}}{\sum_j \sum_{t' < t} b_{tt'} x_{jt'}}.
\end{equation}

Note that for the first time step $t=0$, we cannot access past time steps to calculate spatio-temporal expectations $\mu_{i0}^{(ksw)}$.

We can now simply plug in our new spatio-temporal expectations into the Moran's I metric at time $t$ by replacing 
%$z_{it} = x_{it} - \bar{x}_{t}$ with $z_{it} = x_{it} - \mu_{it}$ 
spatial only expectations with a spatio-temporal expectation of our choice. As such, we define our novel measure of \textbf{spa}tio-\textbf{te}mporal association (SPATE),

\begin{equation} \label{eq:spate}
\begin{split}
    \begin{aligned}
    S_{it} (x, w) = (n_{i}-1) \frac{z_{it}}{\sum^{n_{x}}_{j=1} z_{jt}^2}  \sum^{n_x}_{j=1, j \neq i} w_{i,j} z_{jt}
    \end{aligned}
\end{split}
\end{equation}
where $z_{it} = x_{it} - \mu_{it}$ and $\mu_{it}$ can be any option from $\mu_{it}^{(k)}$, $\mu_{it}^{(kw)}$ and $\mu_{it}^{(ksw)}$. When using $\mu_{it}^{(ksw)}$, SPATE does not return values for $t=0$, as no previous time steps are available to calculate spatio-temporal expectations. See Fig \ref{fig1} for an illustration of the three proposed options.

SPATE measures spatio-temporal autocorrelation at the input resolution. Its behavior can be closely related to that of the Moran's I metric. While Moran's I evaluates the deviance $z_{i}$ between each pixel and the spatial expectation, SPATE does so by using the spatio-temporal expectation $z_{it}$. %uses the spatio-temporal expectation $z_{it}$ to compute the deviance.}
Like Moran's I, SPATE acts as a detector of spatio-temporal clusters and change patterns. Like Moran's I, SPATE identifies positive and negative space-time autocorrelation, i.e. homogeneous areas of similar behavior and outliers that behave differently from their immediate neighborhood. The difference between Moran's I and SPATE is that the later explicitly captures space-time interactions. For example, if pixel $x_i$ and all other data points (not just its neighbors) are increased at a given time step $t$, Moran's I at time $t$ (for all points) will be high, but SPATE will not be. SPATE of $x_{it}$ will be high if \textit{(a)} pixel $x_{it}$ is high compared to its expectation at the same time $t$, \textit{(b)} its neighbors are high compared to their expectations, while \textit{(c)} its non-neighbors are not particularly high compared to their expectations. 

In the $kw$ and $ksw$ settings, the lengthscale parameter $l$ governs whether the metric captures longer or shorter term temporal patterns. For example, if pixel $x_{it}$ and its neighbors increase slowly over time, that change will only cause SPATE to be high for larger lengthscales $l$, while smaller $l$ values imply that current values are compared to those that are close in time. As such, the lengthscale determines what changes are considered "slow" (incorporated into the mean, not detected as space-time interaction) and "fast" (current values are different from the mean, detected as space-time interaction). The $ksw$ setting further allows for scenarios where we might wish to compute the metric based on previous time-steps alone, i.e. to preserve sequential logic.
%\begin{figure*}[h!]%{0.4\textwidth}
%\vskip -0.1in
%\centering
%\includegraphics[width=0.4\textwidth]{figures/emb.png}
%\includegraphics{fig1.pdf}
%\vskip -0.1in
%\caption{The $SPATE$ metric in its different forms, calculated from a sequence of point process intensities from the LGCP datasets. While $SPATE^{k}$ and $SPATE^{kw}$ are visually undistinguishable, $SPATE^{skw}$ changes as increasingly more past time steps become available, converging to $SPATE^{kw}$ at the final time step.}
%\label{fig2}
%end{figure*}

\begin{figure}[h!]
    \centering
    \includegraphics[width=0.481\textwidth]{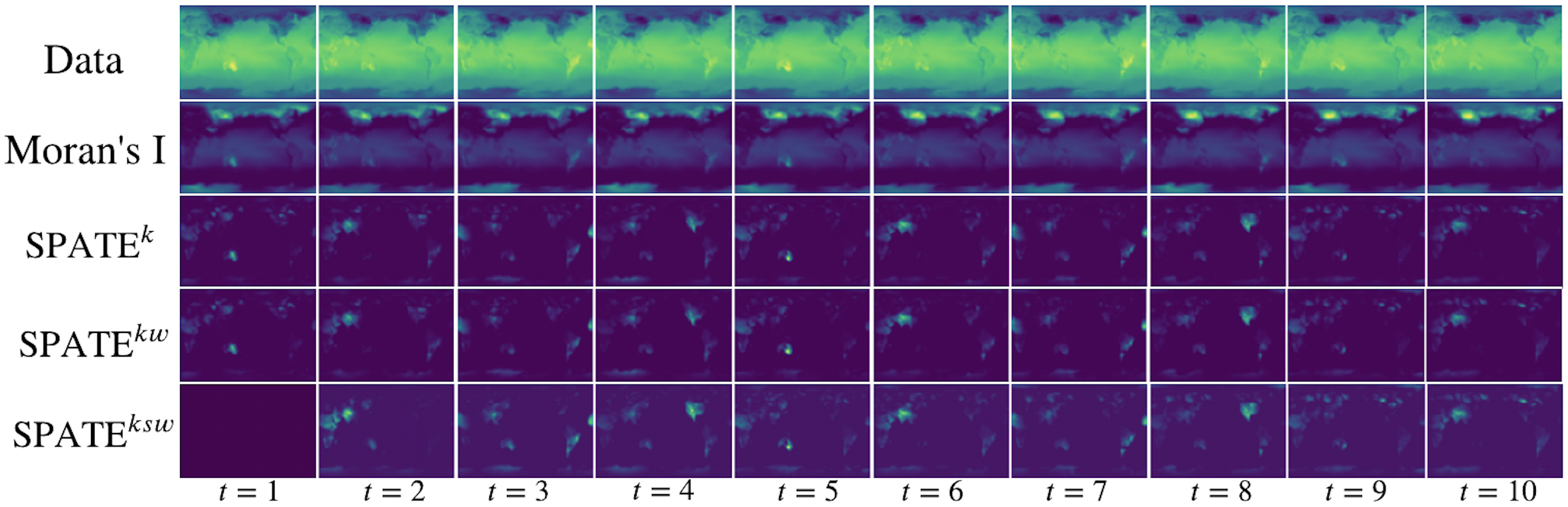}
    \caption{The SPATE metric in its different forms computed for an example from the LGCP datasets.  While $\text{SPATE}^{k}$ and $\text{SPATE}^{kw}$ are visually undistinguishable, $\text{SPATE}^{ksw}$ changes as increasingly more past time steps become available, converging to $\text{SPATE}^{kw}$ at time $T$. We can also observe how the Moran's I metric remains static in time, while all versions of SPATE behave dynamically in space and time.}
    \label{fig:lgcp}
\end{figure} 

\subsection{COT-GAN}
\label{subseq:cot-gan}

Built upon the theory of Causal Optimal Transport (COT), COT-GAN was introduced by \citet{Xu2020} as an adversarial algorithm to train implicit generative models for sequential learning.  COT can be considered as a maximization over the classical (Kantorovich) optimal transport (OT) with a temporal causality constraint, which restricts the transporting of mass on the arrival sequence at any time $t$ to depend on the starting sequence only up to time $t$. This motivated us to design the spatio-temporal expectation $\mu_{it}^{ksw}$ in order to respect the nature of sequential data that are generated in an autoregressive manner.

In sequential generation, we are interested in learning a model that produces $y=(y_1,...,y_T) \in \R^{n\times T}$ to mimic $x=(x_1,...,x_T) \in \R^{n\times T}$.
Given two probability measures 
$\mu,\nu$ defined on $\R^{n\times T}$, and a cost function $c:\R^{n\times T}\times\R^{n\times T}\to\R$, 
the causal optimal transport of $\mu$ into $\nu$ is formulated as
\begin{equation}\label{eq:COT}
\Wcal^\Kcal_c(\mu, \nu) := \inf_{\pi \in \Pi^{\Kcal}(\mu, \nu)} \E^{\pi}[c(x,y)],
\end{equation}
where $\Pi^{\Kcal}(\mu,\nu)$ is the set of probability measures $\pi$ on $\R^{n\times T}\times\R^{n\times T}$ with marginals $\mu,\nu$, which also satisfy the constraint
\begin{equation*} \label{eq:causal_pi}
\pi(dy_t|dx_{1:T})=\pi(dy_t|dx_{1:t})\quad \text{for all\, $t=1,\cdots,T-1$}.
\end{equation*} Such plans in $\Pi^{\Kcal}(\mu,\nu)$ are called \emph{causal transport plans}. Here $c(x,y)$ is a cost function that measures the loss incurred by transporting a unit of mass from $x$ to $y$. $\Wcal^\Kcal_c(\mu,\nu)$ is thus the minimal total cost for moving the mass $\mu$ to $\nu$ in a causal way. 
% Intuitively, the probability mass moved to the target sequence at time $t$ only depends on the starting sequence up to time $t$.

In comparison, the (classic) OT is defined on $\R^{n}$ %rather than the path space $\R^{n \times T}$. Additionally, OT
and differs from COT by searching an optimal plan that leads to the least cost in a less restricted space $\Pi(\mu,\nu)$ which contains all transport plans, 
\begin{equation}\label{eq:OT}
\Wcal_c(\mu, \nu) := \inf_{\pi \in \Pi(\mu, \nu)} \E^{\pi}[c(x,y)].
\end{equation}

COT-GAN constructed an adversarial objective by reformulating the COT \eqref{eq:COT} as below, 
\begin{equation}\label{eq:sup_OT}
\Wcal^\Kcal_{c}(\mu, \nu) = \sup_{c^\Kcal\in\Ccal^\Kcal(\mu,c)} \inf_{\pi \in \Pi(\mu, \nu)} \{ \E^{\pi}[c^\Kcal(x,y)]\},
\end{equation}
where $\Ccal^\Kcal(\mu,c)$ is a special family of costs that encode the causality constraint, see Appendix A for details. Note that the inner optimization problem is equivalent to OT in \eqref{eq:OT} with a specific class of cost function, i.e., $\Wcal_{c^{\Kcal}}(\mu, \nu) = \inf_{\pi \in \Pi(\mu, \nu)} \{ \E^{\pi}[c^\Kcal(x,y)]\}$.  
%where the family of costs $\Ccal^\Kcal(\mu,\nu)$ is given by
%\begin{align*}\label{eq:L_set}
%\Ccal^\Kcal(\mu,\nu) :=\Bigg\{ & c(x,y)+\sum_{j=1}^J \sum_{t=1}^{T-1} h^j_t(y)\Delta_{t+1}M^j(x): \\
%& J\in\N, (h^j,M^j)\in\Hcal(\mu) \Bigg\},
%\end{align*}
%where $\Delta_{t+1}M(x) := M_{t+1}(x_{1:t+1}) - M_t(x_{1:t})$ and $\Hcal(\mu)$ is a set of functions depicting causality:
%\begin{align*}
%\Hcal(\mu):=\{ & (h,M) : h=(h_t)_{t=1}^{T-1},\ h_t\in\Ccal_b(\R^{n\times t}), \\
% &  M=(M_t)_{t=1}^{T}\in\Mcal(\mu),M_t\in \Ccal_b(\R^{n\times t})\},
%\end{align*}
%with $\Mcal(\mu)$ being the set of martingales on $\R^{d\times T}$ w.r.t. the canonical filtration and the measure $\mu$, and  $\Ccal_b(\R^{d\times t})$ the space of continuous, bounded functions on $\R^{d\times t}$. 

In the implementation of COT-GAN, learning is done via stochastic gradient descent (SGD) on mini-batches. Given a distribution $\zeta$ on some latent space $\Zcal$, the generator $g_\theta$ is a function that maps a latent variable $z \sim \zeta$ to the generated sequence $y$ in the path space.  Given a mini-batch of size $m$ from training data $\{x^d_{1:T}\}_{i=1}^m$ and that from generated samples $\{y^d_{1:T}\}_{i=1}^m$, we define the empirical measures for the mini-batches as
\begin{align*}
& \hat{\mu} :=\frac{1}{m}\sum_{d=1}^m\delta_{{x}^d_{1:T}},\quad \hat{\nu}_\theta:=\frac{1}{m}\sum_{d=1}^m\delta_{{y}^d_{1:T}},
\end{align*}
where $\hat{\nu}_\theta$ incorporates the parameterization of $g_\theta$. %The discriminator in COT-GAN is formulated by parameterizing the cost function $c^\Kcal$ by $\varphi$.
%Moreover, the dimensionality of the sets of  $\vh:=(h^j)_{j=1}^J$ and $\vM:=(M^j)_{j=1}^J$ is bounded by a fixed $J\in\mathbb{N}$. The discriminator in COT-GAN is formulated by parameterizing $\vh_{\varphi_1}$ and $\vM_{\varphi_2}$ in the cost function $c^\Kcal$ as two separate neural networks that respect causality,
%\begin{align}
%c^\Kcal_{\varphi}(x,y) = c(x,y)+\sum_{j=1}^J \sum_{t=1}^{T-1} h^j_{\varphi_1, t}(y)\Delta_{t+1}M^j_{\varphi_2}(x),
%\end{align}
%where $\varphi := (\varphi_1, \varphi_2)$ and $J$ corresponds to the output dimensionality of the two networks. Thus, we update the parameters based upon the loss given by \eqref{eq:sup_OT} between the empirical distributions of two mini-batches, 
%Thus, we update the parameters based upon the loss given by \eqref{eq:sup_OT} between the empirical distributions of two mini-batches, 
Thus, the duality of COT given by \eqref{eq:sup_OT} between the empirical measures of two mini-batches can be written as
\begin{equation}\label{eq:sup_OT_mini}
\Wcal^\Kcal_{c}(\hat{\mu}, \hat{\nu}_\theta) = \sup_{c^{\Kcal} \in \Ccal ^{\Kcal}(\hat{\mu}, c) } \Wcal_{c^\Kcal}(\hat{\mu}, \hat{\nu}_\theta).
\end{equation}

Computing \eqref{eq:sup_OT_mini} involves solving the optimization problem of classic OT.  COT-GAN opted for the Sinkhorn algorithm, see \cite{cuturi2013sinkhorn, GPC}, and a modified version of Sinkhorn divergence for an approximation.  %for solving an auxiliary task by adding an entropic regularization to the original transport problems. 
Here we have the entropic-regularized OT defined as
\begin{align*}
\Wcal_{c,\eps}(\mu, \nu) := \inf_{\pi \in \Pi(\mu, \nu)} \{ \E^{\pi}[c(x,y)] - \eps H(\pi)\},\quad \eps>0,
\end{align*}
where $H(\pi)$ is the Shannon entropy of $\pi$. 
%To correct the fact that $\Wcal_{c,\eps}(\mu, \mu) \neq 0$ due to the entropic term, \citet{GPC} proposed the Sinkhorn divergence, defined as
%\begin{equation*}\label{Sink}
%\widehat{\Wcal}_{c,\eps}(\mu, \nu):= 
%2\Wcal_{c,\eps}(\mu, \nu)-\Wcal_{c,\eps}(\mu, \mu)-\Wcal_{c,\eps}(\nu, \nu).
%\end{equation*}
Whilst the Sinkhorn divergence was also computed between two mini-batches, the authors of COT-GAN noticed the bias was better reduced by %using four mini-batches instead. For this reason, they introduced 
the mixed Sinkhorn divergence,
\begin{align*}\label{mixSink}
\widehat{\Wcal}_{c,\eps}^{\text{mix}}(\hat{\mu}, \hat{\nu}, \hat{\mu}', \hat{\nu}') &:= 
\Wcal_{c,\eps}(\hat{\mu}, \hat{\nu}) + \Wcal_{c,\eps}(\hat{\mu}', \hat{\nu}') \\
& \quad -\Wcal_{c,\eps}(\hat{\mu}, \hat{\mu}')-\Wcal_{c,\eps}(\hat{\nu}, \hat{\nu}'),
\end{align*}
where $\hat{\mu}$ and $\hat{\mu}'$ represent the empirical measures corresponding to two different mini-batches of the training data, and $\hat{\nu}$ and $\hat{\nu}'$ are the ones to the generated samples.  
%As the last piece of the puzzle, \citet{Xu2020} enforced $\vM$ to be close to a martingale by a regularization term to penalize deviations from being a martingale on the mini-batches.
%\[
%{p}_{\vM}(\widehat{\mu}):=\frac{1}{mT}\sum_{j=1}^J\sum_{t=1}^{T-1}\Bigg|\sum_{d=1}^m \frac{M^j_{t+1}(x^d_{1:t+1}) - M^j_t(x^d_{1:t})}{\sqrt{\text{Var}[M^j]} + \eta}\Bigg|,
%\]
%where $\text{Var}[M]$ is the empirical variance of $M$ over time and batch, and $\eta>0$ is a small constant.

Finally,  we have the following objective function for COT-GAN:
\begin{equation*}\label{eq:objective}
\inf_{\theta} \sup_{\varphi} \bigg\{ \widehat{\Wcal}_{c^\Kcal_\varphi,\eps}^{\text{mix}}(\hat{\mu}, \hat{\nu}_\theta, \hat{\mu}', \hat{\nu}'_\theta)
- \lambda \big[ p_{{\bf M}_{\varphi_2}} (\hat{\mu}) + p_{{\bf M}_{\varphi_2}} (\hat{\mu}') \big] \bigg\},
\end{equation*}
where the role of discriminator is played by the cost function $c^\Kcal$ parameterized by $\varphi$, and 
$p_{{\bf M}_{\varphi_2}}$ is the martingale penalization required for the formulation of the new class of costs $\Ccal^\Kcal_{\varphi}$, and $\varphi_2$ is a subset of the discriminator parameters $\varphi$, see details in Appendix A. Similar to common GAN training scheme, the discriminator learns the worst-case cost $c^\Kcal_{\varphi}$ by maximizing the objective over $\varphi$, and the generator is updated by minimizing it over $\theta$.

\subsection{SPATE-GAN}

In SPATE-GAN, we integrate our newly devised spatio-temporal metric into the COT-GAN objective function. %Given a distribution $\zeta$ on some latent space $\Zcal$, the generator $g_\theta$ is a function that maps a latent variable $z \sim \zeta$ to the generated sequence $y$ in the path space.  
We compute the embedding for each $x^d_{it}$ and $y^d_{it}$ in minibatches $\{x^d_{1:T}\}_{d=1}^m$ and $\{y^d_{1:T}\}_{d=1}^m$ by 
\begin{align*}
\hat{x}^d_{it} = S_{it}(x^d_{1:T}, w) \quad \text{and} \quad
\hat{y}^d_{it} = S_{it}(y^d_{1:T}, w),
\end{align*}
where the binary spatial weight matrix $w$ is pre-defined.

The corresponding embeddings are then concatenated with the training data and generated samples on the channel dimension. We define the empirical measures for the concatenated sequences as
\begin{align*}
& \widehat\mu^{e}:=\frac{1}{m}\sum_{d=1}^m\delta_{\text{concat}({x}^d_{1:T},\hat{x}_{1:T}^d))},\\
& \widehat\nu^{e}_\theta:=\frac{1}{m}\sum_{d=1}^m\delta_{ \text{concat}({y}^d_{1:T},\hat{y}_{1:T}^d)},
\end{align*}
where $\text{concat}(., .)$ is an operator that concatenates inputs along the channel dimension.

We thus arrive at the objective function for SPATE-GAN:
\begin{align*}\label{eq:objective}
\inf_{\theta} \sup_{\varphi} & \bigg\{ \widehat{\Wcal}_{c^\Kcal_\varphi,\eps}^{\text{mix}}(\widehat\mu^{e}, \widehat\nu^{e}_\theta, \widehat\mu^{e'}, \widehat\nu^{e'}_\theta) \\
& \quad - \lambda \big[ p_{{\bf M}_{\varphi_2}} (\widehat\mu^{e}) + p_{{\bf M}_{\varphi_2}} (\widehat\mu^{e'}) \big] \bigg\}.
\end{align*}
We maximize the objective function over $\varphi$ to search for a worst-case distance between the two empirical measures, and minimize it over $\theta$ to learn a distribution that is as close as possible to the real distribution.
The algorithm is summarized in Algorithm \ref{alg}. Its time complexity scales as $\mathcal{O}((J+2n)LTm^2)$ in each iteration where $J$ is the output dimension of the discriminator (see Appendix A for details), and $L$ is the number of Sinkhorn iterations (see \citet{GPC, cuturi2013sinkhorn} for details).

In the experiment section, we will compare SPATE-GAN with three different expectations $\mu_{it}^{(k)}$, $\mu_{it}^{(kw)}$ and $\mu_{it}^{(ksw)}$ in the computation of SPATE. Hence, we denote the corresponding models as $\text{SPATE-GAN}^k$, $\text{SPATE-GAN}^{kw}$, and $\text{SPATE-GAN}^{ksw}$, respectively.

Last, we would like to emphasize that, although all three embeddings consider the space-time interactions in a certain way, the non-anticipative assumption of $\mu_{it}^{(ksw)}$ is consistent with the generating process of the type of data we are investigating.  As the causality constraint in COT-GAN also restricts the search of transport plans to those that satisfy non-anticipative transporting of mass, $\text{SPATE-GAN}^{ksw}$ is a model that fully respects temporal causality in learning, whilst $\text{SPATE-GAN}^{k}$ and $\text{SPATE-GAN}^{kw}$ also combine information from the future.  
\begin{algorithm}[t]
\SetAlgoLined
\KwData{$\{{x}^d_{1:T}\}_{d=1}^{n}$ (input data), $\zeta$ (latent distribution)}
\KwParam{$\theta_0$, $\varphi_0$ (parameter initializations), $m$ (batch size), $\eps$ (regularization parameter),  %$L$ (number of Sinkhorn iterations), 
$\alpha$ (learning rate), $\lambda$ (martingale penalty coefficient)}
 Initialize: $\theta \leftarrow \theta_0$, $\varphi \leftarrow \varphi_{0}$\\
 \For{$b = 1, 2,\dots$}{
    ~Sample $\{{x}^d_{1:T}\}_{d=1}^{m}$ from real data;\\[0.1cm]
    %~Learn features from input sequence $\{{x}^i_{1:T}\}_{i=1}^{m}$ and get the last state which contains all past information from encoder:
    %~$\{{f}^i_{T}\}_{i=1}^{m} = g^{e}_{\theta} (\{ {x}^i_{1:T}\}_{i=1}^{m})$;\\[0.1cm]
    ~Sample $\{{z}^d_{1:T}\}_{d=1}^{m}$ from 
    ~$\zeta$;\\[0.1cm]
    ~Generate sequences from latent: $(y_{1:T}^d) \leftarrow g_{\theta}({z}^d_{1:T})$;\\[0.1cm]
    ~Compute the embeddings: $\hat{x}^d_{it} = S_{it}(x^d_{1:T}, w), \quad \hat{y}^i_{it} = S_{it}(y^i_{1:T}, w)$;\\[0.1cm]
    ~Concatenated the data with embeddings: $\text{concat}({x}^d_{1:T},\hat{x}_{1:T}^d), \text{concat}({y}^d_{1:T},\hat{y}_{1:T}^d)$ ;\\[0.1cm]
    ~Update discriminator parameter: \\ [0.1cm]
    ${\varphi} \leftarrow {\varphi} + \alpha \nabla_\varphi \big( \widehat{\Wcal}_{c^\Kcal_\varphi,\eps}^{\text{mix}}(\widehat\mu^{e}, \widehat\nu^{e}_\theta, \widehat\mu^{e'}, \widehat\nu^{e'}_\theta) - \lambda \big[ p_{{\bf M}_{\varphi_2}} (\widehat\mu^{e}) + p_{{\bf M}_{\varphi_2}} (\widehat\mu^{e'}) \big] \big)$; \\[0.1cm]
    ~Sample $\{{z}^d_{1:T}\}_{d=1}^{m}$ from  $\zeta$;\\[0.1cm]
    ~Generate sequences from latent: $(y_{1:T}^d) \leftarrow g_{\theta}({z}^d_{1:T})$;\\[0.1cm]
    ~Compute the embeddings: $\hat{x}^d_{it} = S_{it}(x^d_{1:T}, w), \quad \hat{y}^d_{it} = S_{it}(y^d_{1:T}, w)$;\\[0.1cm]
    ~Concatenated the data with embeddings: $\text{concat}({x}^d_{1:T},\hat{x}_{1:T}^d), \text{concat}({y}^d_{1:T},\hat{y}_{1:T}^d)$ ;\\[0.1cm]
    ~Update generator parameter: $\theta \leftarrow \theta - \alpha \nabla_\theta\left( \widehat{\Wcal}_{c^\Kcal_\varphi,\eps}^{\text{mix}}(\widehat\mu^{e}, \widehat\nu^{e}_\theta, \widehat\mu^{e'}, \widehat\nu^{e'}_\theta) \right)$;
}
 \caption{training SPATE-GAN by SGD}
 \label{alg}
\end{algorithm}

%% file: experiments.tex
To empirically evaluate SPATE-GAN, we use three datasets characterized by different spatio-temporal complexities.

\paragraph{Extreme Weather (EW)} This dataset, introduced by \citet{Racah2017}, was originally proposed for detecting extreme weather events from a range of climate variables (e.g. zonal winds, radiation). Each of these climate variables is observed four times a day for a $128 \times 192$ pixel representation of the whole earth. We chose to model surface temperature%, as this variable 
as it comes with several interesting spatio-temporal characteristics: It exhibits both static (e.g. continent outlines) and dynamic patterns as well as abnormal patterns (e.g. in the presence of tropical cyclones or atmospheric rivers). Furthermore, simulating climate data is an important potential downstream application of deep generative models.

\paragraph{LGCP} This dataset represents the intensities (number of events in a grid cell) of a log-Gaussian Cox process (LGCP), a continuous spatio-temporal point process. LGCPs are a popular class of models for simulating contagious spatio-temporal patterns and have various applications, for example in epidemiology. We simulate $300$ different LGCP intensities on a $64 \times 64$ grid over $10$ time steps using the \textit{R} package \textit{LGCP} \cite{Taylor2015}.

\paragraph{Turbulent Flows (TF)} This dataset, proposed by \cite{Wang2020}, simulates velocity fields according to the Navier-Stokes equation. This is a class of partial differential equations describing the motion of fluids. Fluid dynamics and simulation is another potential application of deep generative models. Following the approach of \citet{Wang2020}, we divide the data into $7$ steps of $64 \times 64$ pixel frames. Please note that we only utilize the first velocity field, so that all our utilized datasets are single-channel.

\subsection{Baselines and evaluation metrics}

We use COT-GAN \cite{Xu2020} and GAN proposed by \cite{GPC}, which we name as SinkGAN, as base models. We augment both models with our new embedding loss, using SPATE with $k$, $kw$ and $ksw$ configurations. We refer to all models using a COT-GAN backbone in combination with our new embedding loss as SPATE-GAN. We further denote the SinkGAN models corresponding to three SPATE settings as $\text{SinkGAN}^{k}$, $\text{SinkGAN}^{kw}$ and $\text{SinkGAN}^{ksw}$.
To compare our approach to a non-time-sensitive embedding, we also deploy models using the Moran's I metric using the same embedding loss procedure, %. These baselines are 
denoted as $\text{COT-GAN}^{M}$ and $\text{SinkGAN}^{M}$. 

To compare our GAN output to real data samples, we use three different metrics: Earth Mover Distance (EMD), Maximum Mean Discrepancy (MMD) \cite{Borgwardt2006} and a classifier two-sample test based on a k-nearest-neighbor (KNN) classifier with $k=1$ \cite{Lopez-Paz2019}. All these measures are general purpose GAN metrics. While GAN metrics specialized on video data exist, they rely on extracting features from models 
% such as VGG 16. This feature extraction requires 
pre-trained on three-channel RGB video data. As we are working with single-channel, non-image data, these methods are not applicable in our case.

%To compute our three metrics, let us first assume that we have a set of real data samples ($\mathcal{P}$) and synthetic data samples ($\mathcal{S}$).
%EMD is defined as:
%\begin{equation}
%    EMD(\mathcal{P},\mathcal{S}) = \min_{\phi: \mathcal{P} \rightarrow \mathcal{S}} \sum_{p \in \mathcal{P}} \| p - \phi (p)\|
%\end{equation}
%where $\phi: \mathcal{P} \rightarrow \mathcal{S}$ is a bijection.
%MMD is defined as:
%\begin{equation}
%\begin{aligned}
%    \widehat{MMD}^{2}(\mathcal{P},\mathcal{S}) = 1 / n(n - 1) \sum k(p,p) +  \\
%    1 / n(n - 1) \sum k(s,s) -  2 / n^{2} \sum k(p,s) 
%\end{aligned}
%\end{equation}
%where $k$ denotes a positive-definite kernel (e.g. RBF kernel) and $n$ is the number of (real or synthetic) samples.

%Lastly, to compute the KNN score, we first split our real and synthetic samples $\mathcal{P}$ and $\mathcal{S}$ into training and test datasets $\mathcal{D}_{tr}$ and $\mathcal{D}_{te}$ so that $\mathcal{D} = \mathcal{D}_{tr} \cup \mathcal{D}_{te}$. We train the KNN classifier $f: \mathcal{X}_{tr} \rightarrow [0,1]$ using training data. The accuracy of the trained classifier is then obtained using test samples $\mathcal{D}_{te}$ and given as:
%\begin{equation}
%    \hat{t} = 1/n_{te} \sum_{(z_{i},l_{i}) \in \mathcal{D}_{te}} \mathbb{I} [(f(z_{i}) > 1/2) = l_{i}]
%\end{equation}
%where $f(z_{i})$ estimates the conditional probablility distribution $p(l=1|z_{i})$. A classifier accuracy approaching random chance (50\%) indicates better synthetic data. As suggested by \citet{Lopez-Paz2019}, we use a $1$-NN classifier to obtain the score.

\subsection{Experimental Setting}

We compare SPATE-GAN to a range of baseline configurations. We use the same GAN architecture for all these settings to ensure comparability. 
Our GAN generators feed the noise input through two LSTM layers to obtain time-dependent features. These are then mapped into the desired shape for deconvolutional operations using a fully-connected layer with a leaky ReLU activation. Lastly, 4 deconvolutional layers map the output into video frames, all also with leaky ReLU activations. 
Our discriminators initially feed video input through three convolutional layers with leaky ReLU activations. The outputs from the convolutional operations are then reshaped and fed through two LSTM layers to create the final discriminator outputs. 

All our models are implemented in PyTorch \cite{Paszke2019} and optimized using the Adam algorithm \cite{Kingma2015}. Our experiments are conducted on a single Geforce 1080Ti or RTX 3090 GPU.  Further training details can be found in Appendix B.  

\subsection{Results}
\begin{figure}[h]
    \centering
    \includegraphics[width=0.49\textwidth]{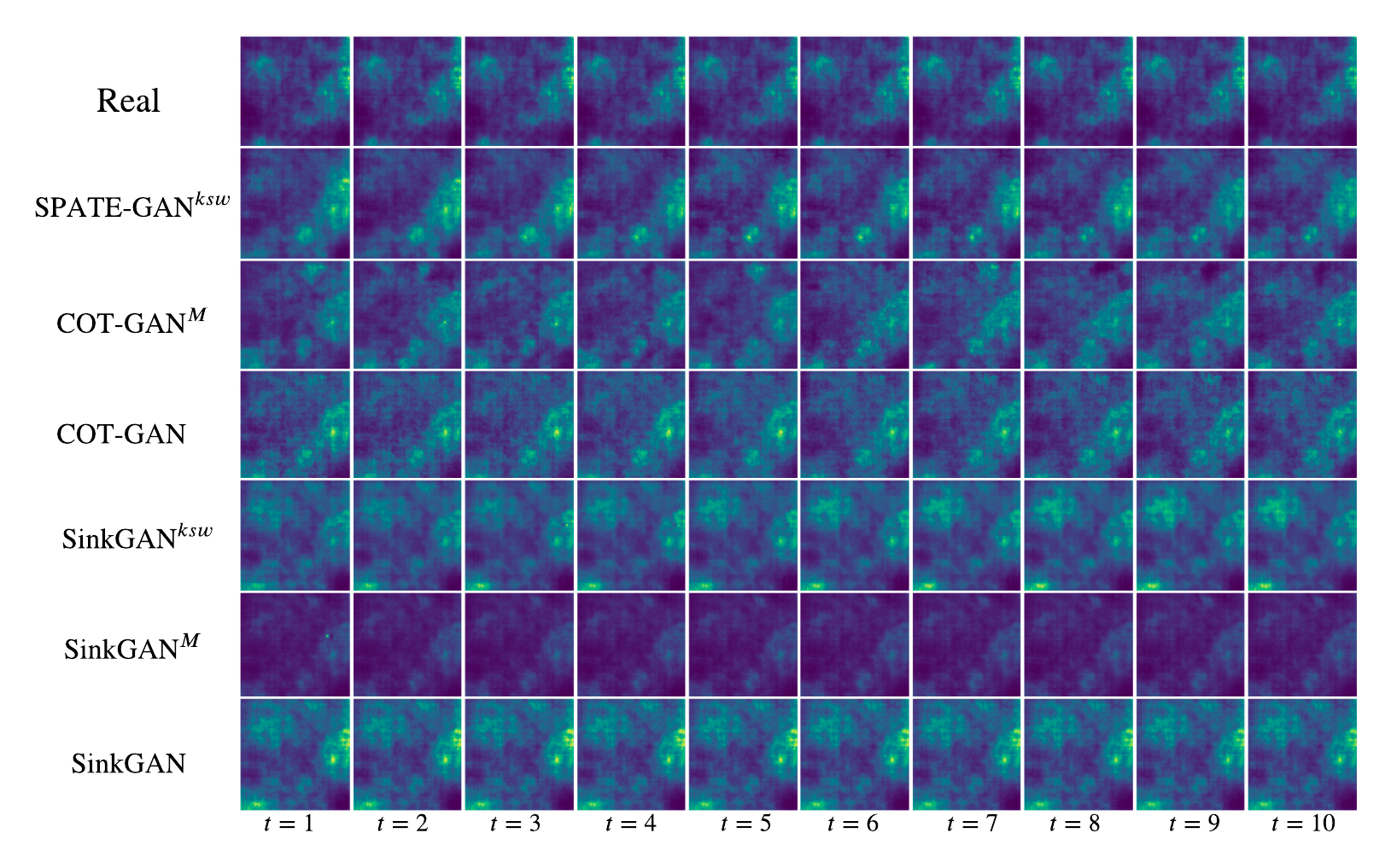}
    \caption{Selected samples for LGCP dataset.}
    \label{fig:lgcp}
\end{figure}

\begin{figure}[h!]
    \centering
    \includegraphics[width=0.49\textwidth]{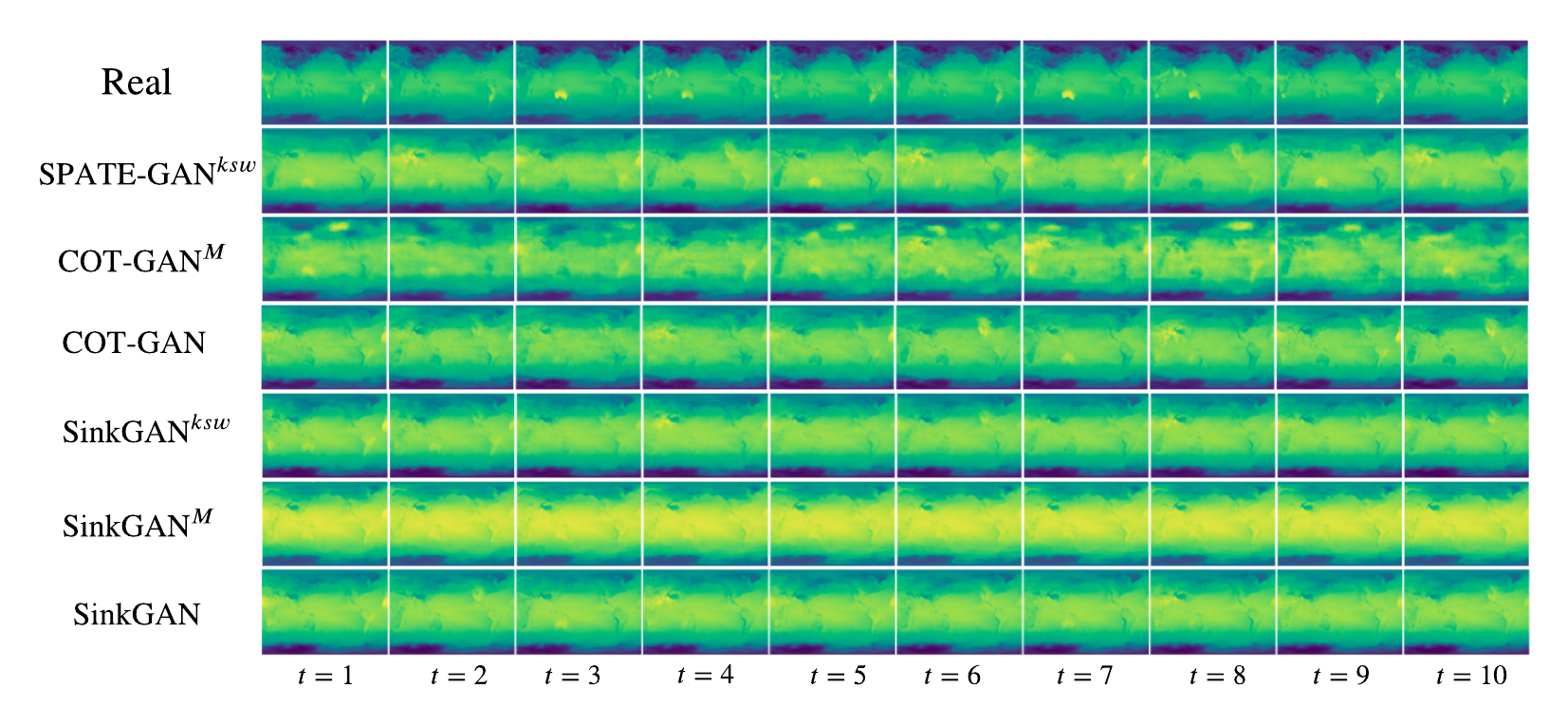}
    \caption{Selected samples for Extreme Weather dataset.}
    \label{fig:eweather}
\end{figure}

Results from our experiments are shown in Table \ref{tab:compare1}. Visual comparisons between real and generated data from the different models are shown in Figures \ref{fig:lgcp},\ref{fig:eweather}, and \ref{fig:tf}. For larger figures including results from all tested model configurations, please see the Appendix D. Through all experiments we can observe that $\text{SPATE-GAN}^{ksw}$ consistently outperforms the competing approaches, achieving the best scores across all datasets and evaluation metrics.

\begin{table}[!h]
\centering
\caption{Evaluations for LGCP, EW and TF datasets. Lower values in EMD and MMD indicate better sample quality, while values close to 0.5 are more desirable for KNN.}
% \begin{tabular}[!t]{lccccc}
\resizebox{8.4cm}{!}{%
\begin{tabular}[!t]{lcccccc}
\hline
\hline
\bf{LGCP} &  EMD  &  MMD  &  KNN  \\
\hline
SinkGAN    & 12.46 (0.02)   &  0.38 (0.001)   &  0.14 (0.001) \\
$\text{SinkGAN}^M$     & 12.46 (0.02)   &  0.38 (0.001)  & 0.14 (0.001)  \\
$\text{SinkGAN}^{k}$   &  12.65 (0.03)  &  0.38 (0.001)  & 0.15 (0.001) \\
$\text{SinkGAN}^{kw}$    &   10.60 (0.01)  &   0.63 (0.008)  & 0.30 (0.002)   \\
$\text{SinkGAN}^{ksw}$   &  13.33 (0.01)  &  0.36 (0.001)  &  0.38 (0.003) \\
\hline
COT-GAN          &  12.38 (0.02) &  \bf{0.30 (0.001)}  &  0.20 (0.004) \\
$\text{COT-GAN}^M$   &  12.38 (0.02)   & \bf{0.30 (0.001)}   &  0.20 (0.004) \\
$\text{SPATE-GAN}^k$          &  11.56 (0.02) &  0.32 (0.01) & 0.31 (0.01) \\
$\text{SPATE-GAN}^{kw}$   &  10.92 (0.03) &  0.64 (0.035)   &  0.15 (0.006)  \\
$\text{SPATE-GAN}^{ksw}$     &  \bf{10.47 (0.02)}  &  \bf{0.30 (0.001)}  &  \bf{0.39 (0.01)} \\
\hline
\hline
\multicolumn{2}{l}{\bf{Extreme Weather}}   &     &     \\
\hline
SinkGAN    &  29.40 (0.05)  &  0.49 (0.001)  &  0.41 (0.004) \\
$\text{SinkGAN}^M$   & 29.27 (0.05)  &  0.72 (0.002)  & 0.22 (0.01)  \\
$\text{SinkGAN}^{k}$   &  32.57 (0.03)   &  0.81 (0.001)   &  0.16 (0.004) \\
$\text{SinkGAN}^{kw}$   &  32.78 (0.05) &  0.81 (0.001) & 0.18 (0.004) \\
$\text{SinkGAN}^{ksw}$  &  30.00 (0.04)  &  0.50 (0.001)   &  0.41 (0.004) \\
\hline
COT-GAN       &  26.66 (0.09)  & 0.43 (0.002)  &  \bf{0.42 (0.002)} \\
$\text{COT-GAN}^M$    &  36.42 (0.14)   &  0.65 (0.002)  &  0.09 (0.01) \\
$\text{SPATE-GAN}^k$    &  33.58 (0.07)  &  0.73 (0.002)  &  0.15 (0.01)\\
$\text{SPATE-GAN}^{kw}$  &  33.36 (0.09)  & 0.72 (0.002) &  0.13 (0.003)  \\
$\text{SPATE-GAN}^{ksw}$ &  \bf{26.24 (0.07)} &  \bf{0.42 (0.002)}  &  \bf{0.42 (0.002)} \\
\hline

\hline
\hline
\multicolumn{2}{l}{\bf{Turbulent Flows}}     &     &     \\
\hline
SinkGAN    &  26.52 (0.007)   &  1.23 (0.001)  &   0.15 (0.001) \\
$\text{SinkGAN}^M$    &  28.02 (0.005)   &  1.22 (0.0002)  & 0.01 (0.002)   \\
$\text{SinkGAN}^{k}$   &   28.14 (0.002)   &    1.32 (0.002)  & 0.08 (0.001) \\
$\text{SinkGAN}^{kw}$    &  30.98 (0.001)  &    1.50 (0.001)&  0.03 (0.001) \\
$\text{SinkGAN}^{ksw}$    &  25.47 (0.008)   &  1.24 (0.0002)  &  0.13 (0.002)   \\
\hline
COT-GAN  &  27.03 (0.01) &  1.22 (0.001)   & \bf{0.16 (0.002)} \\
$\text{COT-GAN}^M$   &  24.93 (0.01)   &1.19 (0.001)  &  0.09 (0.002)  \\
$\text{SPATE-GAN}^k$   &  25.70 (0.02)   &     1.21 (0.001) &  0.12 (0.003)  \\
$\text{SPATE-GAN}^{kw}$   &  24.30 (0.002)  &  1.42 (0.001)  &  0.13 (0.004) \\
$\text{SPATE-GAN}^{ksw}$    &  \bf{22.98 (0.01)}   & \bf{1.16 (0.001)}    &  \bf{0.16 (0.002)}  \\
\hline
\hline
\end{tabular}}
\label{tab:compare1}
\end{table}

This finding is interesting as the $ksw$ setting theoretically looses information over the $k$ and $kw$ approaches, which both have access to future time steps when calculating the SPATE metric. Nevertheless, this result underlines the strong synergies between $\text{SPATE}^{ksw}$ and the COT-GAN backbone: The metric is calculated in sequential fashion and thus respects the same causality constraints that restrict COT-GAN. As such, the outcome, while noteworthy, is not surprising.

This result is strengthened by a comparison with the SinkGAN-based approaches: SinkGAN does not follow the same restrictions and, as we observe, is not improved as consistently by the SPATE-based embedding losses. In fact, in some cases the naive SinkGAN performs better than its derivatives using SPATE or Moran's I based embedding losses.
\begin{figure}[h]
    \centering
    \includegraphics[width=0.38\textwidth]{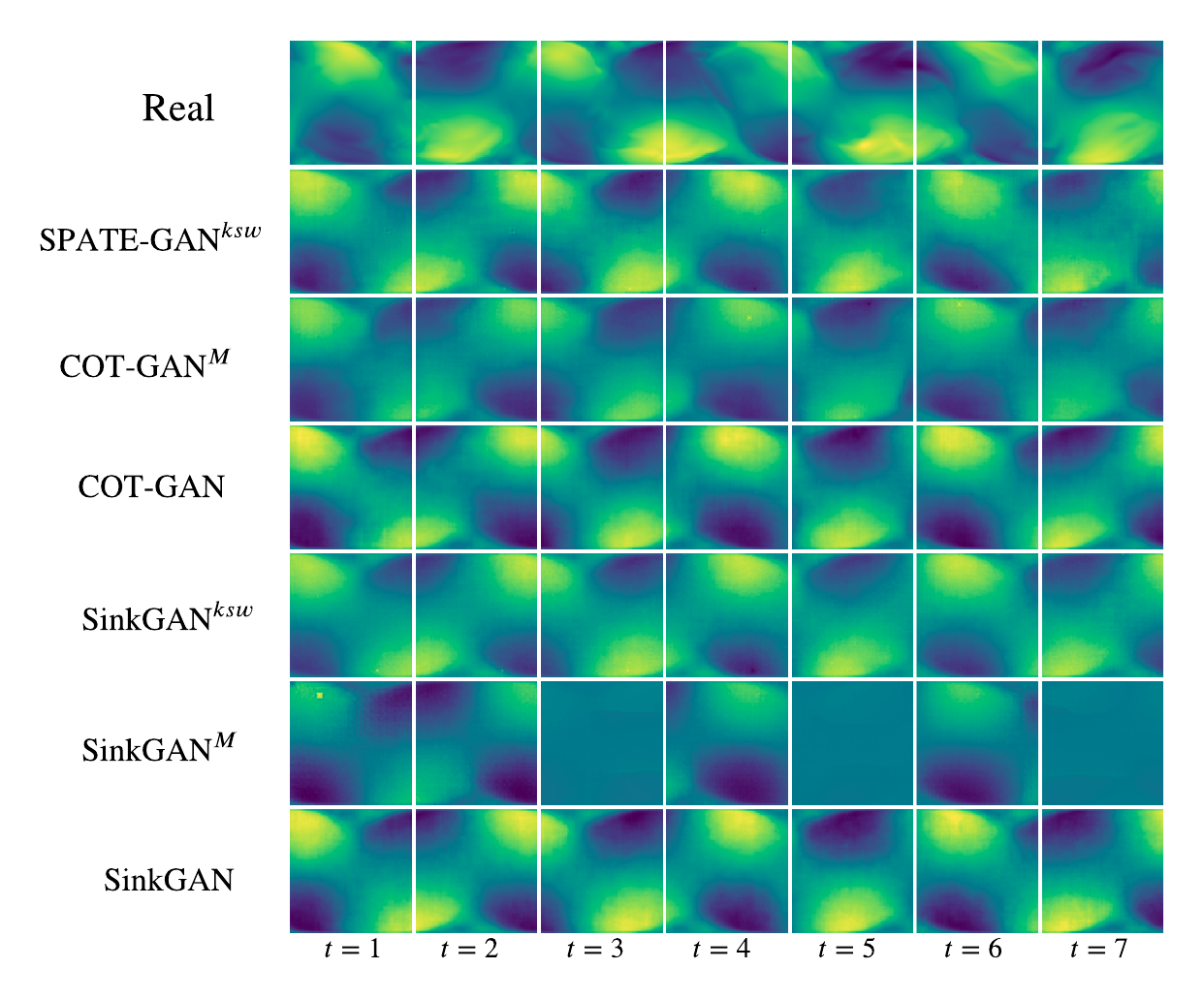}
    \caption{Selected samples for Turbulent Flows dataset.}
    \label{fig:tf}
\end{figure} 

We also observe that throughout all settings, models using Moran's I perform similarly to their naive counterparts. This confirms that in fact, simply using measures of spatial autocorrelation computed over a sequence is not sufficient for capturing complex spatio-temporal effects. On the contrary, the other two SPATE settings, $k$ and $kw$, both appear to have beneficial effects and improve performance.

In summary, our results highlight how COT-GAN combined with a non-anticipative measure of space-time association can improve the modeling of complex spatio-temporal patterns. This finding represents another step on the way towards deep learning methods specialized on the dynamics driving many systems on our planet.

\begin{figure}[h!]
    \centering
    \includegraphics[width=0.48\textwidth]{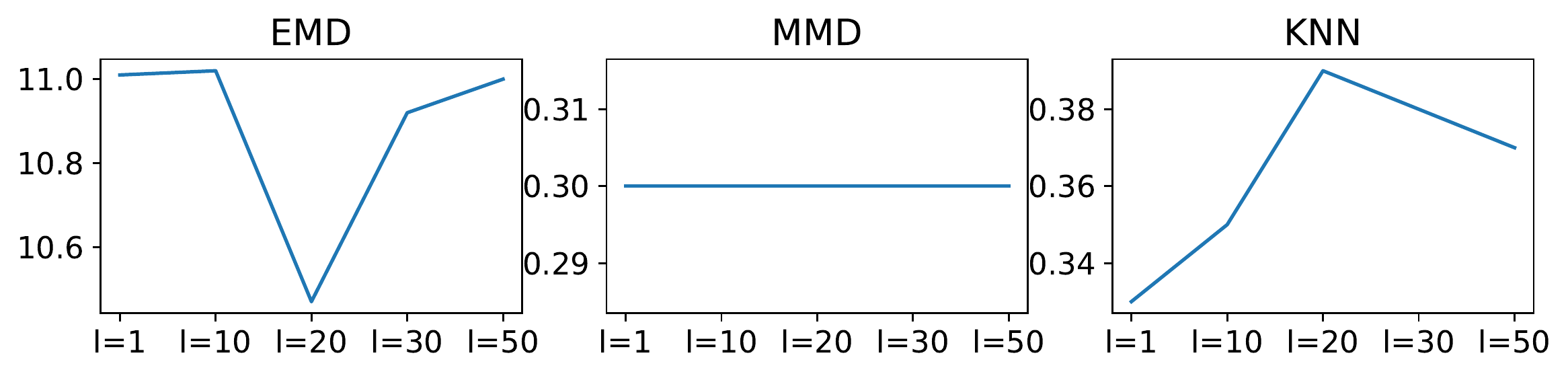}
    \caption{Evaluations of $\text{SPATE-GAN}^{ksw}$ (left: EMD, middle: MMD, and right: KNN) on LGCP dataset given lengthscale $l \in \{1, 10, 20, 30, 50\}$. }
    \label{fig:l}
\end{figure}

Furthermore, we provide an investigation on the impact of the lengthscale parameter $l$ in the spatio-temporal expectations for $l \in \{1, 10, 20, 30, 50 \}$.  As shown in Figure \ref{fig:l},  $l=20$ leads to better EMD and KNN results whilst all MMD scores remain unchanged. For the results presented in this paper, we set $l=20$ in all our experiments.

%% file: conclusion.tex
Recent studies have called for more research into improving deep learning models for spatio-temporal earth systems data \cite{Reichstein2019}. Other academic domains have dealt with these data for many decades and have developed methods for capturing specific spatial and spatio-temporal effects. Inspired by their approaches, we devise SPATE, a measure of spatio-temporal association capable of detecting emerging space-time clusters and homogeneous areas in the data. We then develop a novel embedding loss for video GANs utilizing SPATE as a means of reinforcing the learning of these patterns-of-interest. Our new generative modeling approach, SPATE-GAN, shows performance increases on a range of different datasets emulating the real-world complexities of spatio-temporal dynamics. As such, this study highlights how domain expertise from applied academic areas can help to motivate methodological advances in machine learning.

%% file: appendix.tex
\section*{A: Details for COT-GAN}
The family of cost functions $\Ccal^\Kcal(\mu,c)$ is given by
\begin{align*}
\Ccal^\Kcal(\mu,c) :=\Bigg\{ & c(x,y)+\sum_{j=1}^J \sum_{t=1}^{T-1} h^j_t(y)\Delta_{t+1}M^j(x): \\
& J\in\N, (h^j,M^j)\in\Hcal(\mu) \Bigg\},
\end{align*}
where $\Delta_{t+1}M(x) := M_{t+1}(x_{1:t+1}) - M_t(x_{1:t})$ and $\Hcal(\mu)$ is a set of functions depicting causality:
\begin{align*}
\Hcal(\mu):=\{ & (h,M) : h=(h_t)_{t=1}^{T-1},\ h_t\in\Ccal_b(\R^{n\times t}), \\
 &  M=(M_t)_{t=1}^{T}\in\Mcal(\mu),M_t\in \Ccal_b(\R^{n\times t})\},
\end{align*}
with $\Mcal(\mu)$ being the set of martingales on $\R^{n\times T}$ w.r.t. the canonical filtration and the measure $\mu$, and  $\Ccal_b(\R^{n\times t})$ the space of continuous, bounded functions on $\R^{n\times t}$. 

% The discriminator in COT-GAN is formulated by parameterizing the cost function $c^\Kcal$ by $\varphi$.
Moreover, in the implementation of COT-GAN, the dimensionality of the sets of  $\vh:=(h^j)_{j=1}^J$ and $\vM:=(M^j)_{j=1}^J$ is bounded by a fixed $J\in\mathbb{N}$. The discriminator in COT-GAN is formulated by parameterizing $\vh_{\varphi_1}$ and $\vM_{\varphi_2}$ in the cost function $c^\Kcal$ as two separate neural networks that respect causality,
\begin{align}
c^\Kcal_{\varphi}(x,y) = c(x,y)+\sum_{j=1}^J \sum_{t=1}^{T-1} h^j_{\varphi_1, t}(y)\Delta_{t+1}M^j_{\varphi_2}(x),
\end{align}
where $\varphi := (\varphi_1, \varphi_2)$ and $J$ corresponds to the output dimensionality of the two networks. Thus, we update the parameters based upon the loss given by \eqref{eq:sup_OT} between the empirical distributions of two mini-batches, 

Given a mini-batch of size $m$ from training data $\{x^d_{1:T}\}_{i=1}^m$ we define the empirical measure for the mini-batch as
\begin{align*}
& \hat{\mu} :=\frac{1}{m}\sum_{d=1}^m\delta_{{x}^d_{1:T}}.
\end{align*}

As the last piece of the puzzle, \citet{Xu2020} enforced $\vM$ to be close to a martingale by a regularization term to penalize deviations from being a martingale on the level of mini-batches.
\[
{p}_{\vM}(\widehat{\mu}):=\frac{1}{mT}\sum_{j=1}^J\sum_{t=1}^{T-1}\Bigg|\sum_{d=1}^m \frac{M^j_{t+1}(x^d_{1:t+1}) - M^j_t(x^d_{1:t})}{\sqrt{\text{Var}[M^j]} + \eta}\Bigg|,
\]
where $\text{Var}[M]$ is the empirical variance of $M$ over time and batch, and $\eta>0$ is a small constant.

\section*{B: Training details}

\begin{table}
\centering
\caption{Generator architecture.}
\resizebox{8.4cm}{!}{
 \begin{tabular}{||c c||}
 \hline
 Generator & Configuration  \\ 
 \hline\hline
 Input &  $z \sim \mathcal{N}(\mathbf{0}, \mathbf{I})$ \\ 
 \hline
 0 & LSTM(state size = 64), BN \\
 \hline
 1 & LSTM(state size = 128), BN \\
 \hline
 2 & Dense(8*8*256), BN, LeakyReLU \\
 \hline
 3 & reshape to 4D array of shape (m, 8, 8, 256) \\ 
 \hline
 4 & DCONV(N256, K5, S1, P=SAME), BN, LeakyReLU \\ 
 \hline
 5 & DCONV(N128, K5, S2, P=SAME), BN, LeakyReLU \\ 
 \hline
 6 & DCONV(N64, K5, S2, P=SAME), BN, LeakyReLU \\ 
 \hline
 7 & DCONV(N1, K5, S2, P=SAME) \\
 \hline
\end{tabular}}
\label{table:g_structure}
\end{table}

\begin{table}
\caption{Discriminator architecture.}
\centering
\resizebox{8.4cm}{!}{
 \begin{tabular}{||c c||}
 \hline
 Discriminator & Configuration  \\ 
 \hline\hline
 Input &  \\
 \hline
 0 & CONV(N64, K5, S2, P=SAME), BN, LeakyReLU\\
 \hline
 1 & CONV(N128, K5, S2, P=SAME), BN, LeakyReLU\\
 \hline
 2 & CONV(N256, K5, S2, P=SAME), BN, LeakyReLU\\
 \hline
 3 & reshape to 3D array of shape (m, T, -1)\\ 
 \hline
 4 & LSTM(state size = 256), BN \\ 
 \hline
 5 & LSTM(state size = 64) \\ 
 \hline
\end{tabular}}
\label{table:d_structure}
\end{table}

We used a smaller size of model with the same network architectures as COT-GAN to train all three datasets.  The architectures for generator and discriminator are given in Tables \ref{table:g_structure} and \ref{table:d_structure}. 

Hyperparameter settings are as follows:  the Sinkhorn regularizer $\epsilon=0.8$, Sinkhorn iteration $L=100$, the lengthscale $l=20$ and martingale penalty $\lambda=1.5$.  We used Adam optimizer with learning rate $0.0001$, $\beta_1=0.5$ and $\beta_2=0.9$. All models are trained for $60,000$ iterations.  

\section*{C: Evaluation metrics}

To compute our three metrics, let us first assume that we have a set of real data samples ($\mathcal{P}$) and synthetic data samples ($\mathcal{S}$).
EMD is defined as:
\begin{equation}
    EMD(\mathcal{P},\mathcal{S}) = \min_{\phi: \mathcal{P} \rightarrow \mathcal{S}} \sum_{p \in \mathcal{P}} \| p - \phi (p)\|
\end{equation}
where $\phi: \mathcal{P} \rightarrow \mathcal{S}$ is a bijection.
MMD is defined as:
\begin{equation}
\begin{aligned}
    \widehat{MMD}^{2}(\mathcal{P},\mathcal{S}) = \frac{1}{n(n - 1)} \sum k(p,p) +  \\
    \frac{1}{n(n - 1)} \sum k(s,s) -  \frac{2}{n^2} \sum k(p,s) 
\end{aligned}
\end{equation}
where $k$ denotes a positive-definite kernel (e.g. RBF kernel) and $n$ is the number of (real or synthetic) samples.

Lastly, to compute the KNN score, we first split our real and synthetic samples $\mathcal{P}$ and $\mathcal{S}$ into training and test datasets $\mathcal{D}_{tr}$ and $\mathcal{D}_{te}$ so that $\mathcal{D} = \mathcal{D}_{tr} \cup \mathcal{D}_{te}$. We train the KNN classifier $f: \mathcal{X}_{tr} \rightarrow [0,1]$ using training data. The accuracy of the trained classifier is then obtained using test samples $\mathcal{D}_{te}$ and given as:
\begin{equation}
    \hat{t} = \frac{1}{n_{te}} \sum_{(z_{i},l_{i}) \in \mathcal{D}_{te}} \mathbb{I} \left[\left(f(z_{i}) > \frac{1}{2}\right) = l_{i}\right]
\end{equation}
where $f(z_{i})$ estimates the conditional probability distribution $p(l=1|z_{i})$. A classifier accuracy approaching random chance (50\%) indicates better synthetic data. As suggested by \citet{Lopez-Paz2019}, we use a $1$-NN classifier to obtain the score.

\section*{D: More figures}
In this section, we provide more results in larger figures for visual comparisons.  

\begin{figure*}[h!]
    \centering
    \includegraphics[width=0.98\textwidth]{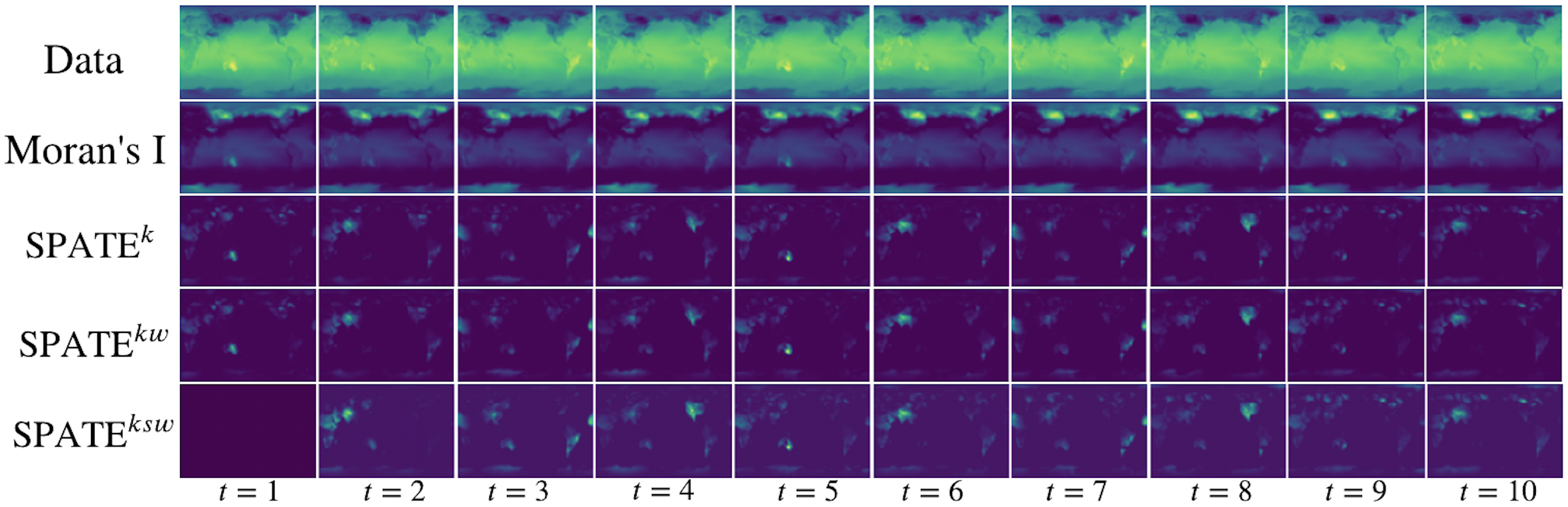}
    \caption{Larger version of Figure 2 for the purpose of visual comparison.}
    \label{fig:eweather_metrics}
\end{figure*} 

\begin{figure*}[h!]
    \centering
    \includegraphics[width=0.98\textwidth]{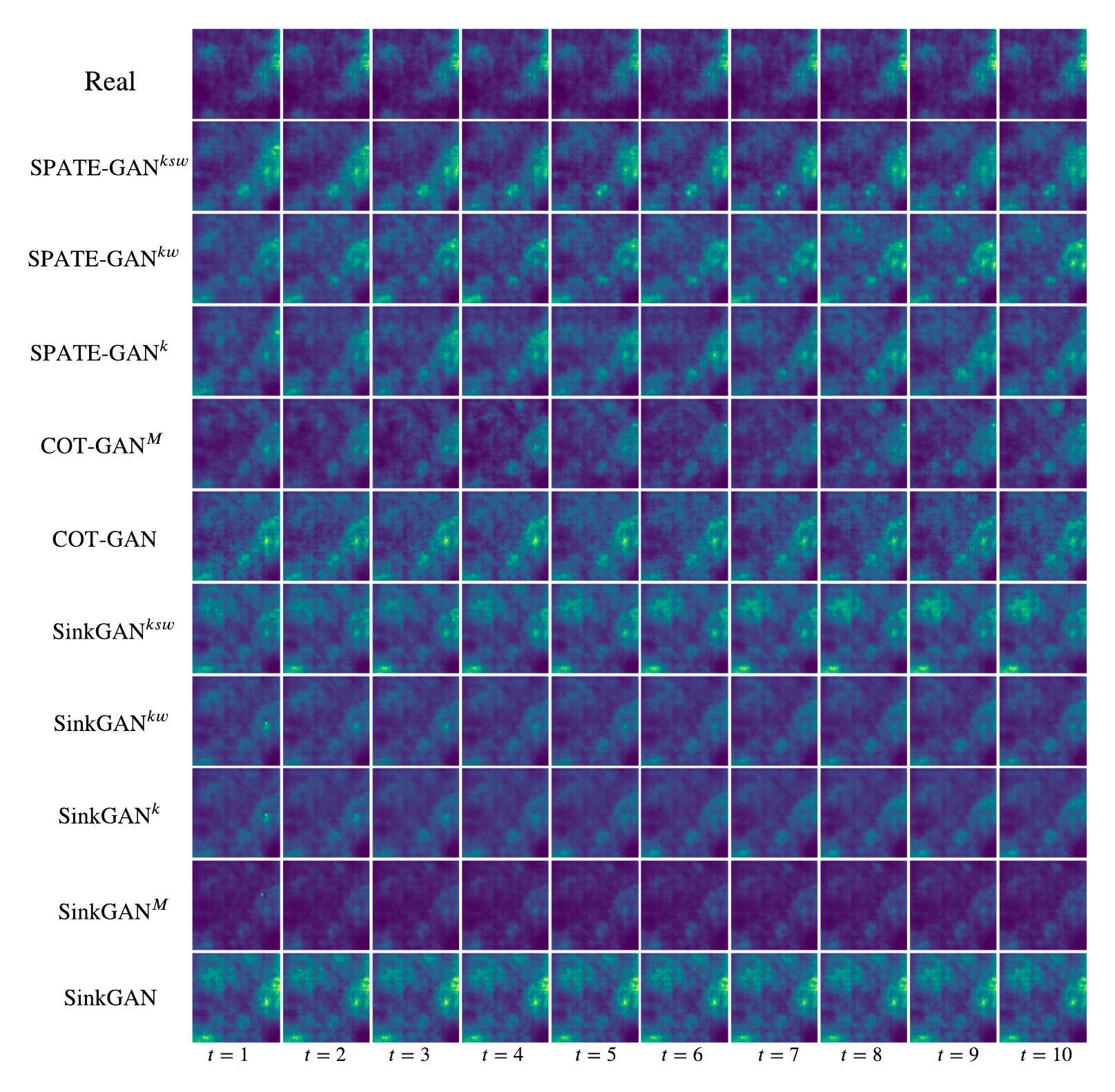}
    \caption{More selected samples for log-Gaussian Cox process (LGCP) dataset.}
    \label{fig:lgcp_full}
\end{figure*} 

\begin{figure*}[h!]
    \centering
    \includegraphics[width=0.98\textwidth]{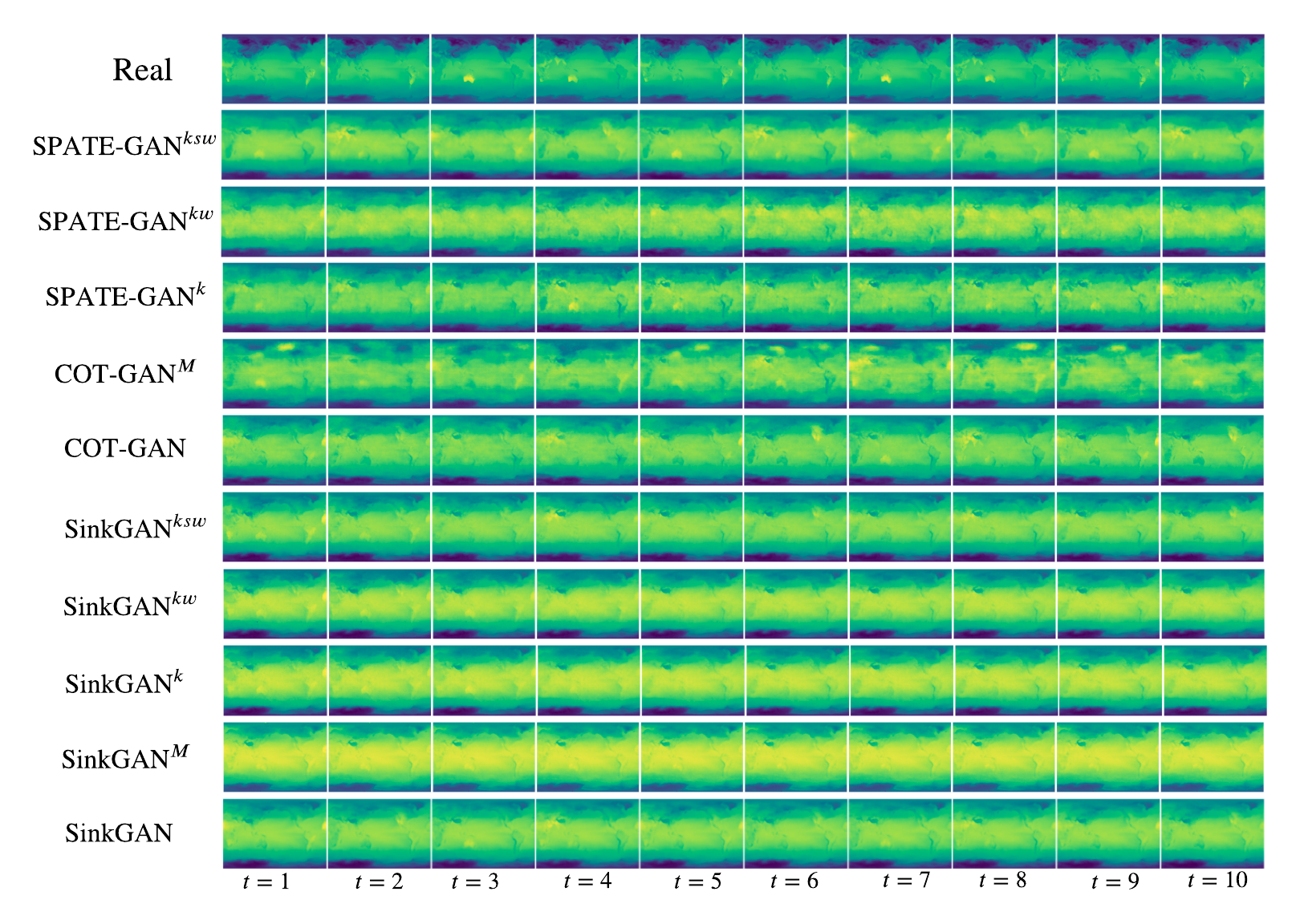}
    \caption{More selected samples for extreme weather (EW) dataset.}
    \label{fig:eweather_full}
\end{figure*} 

\begin{figure*}[h!]
    \centering
    \includegraphics[width=0.98\textwidth]{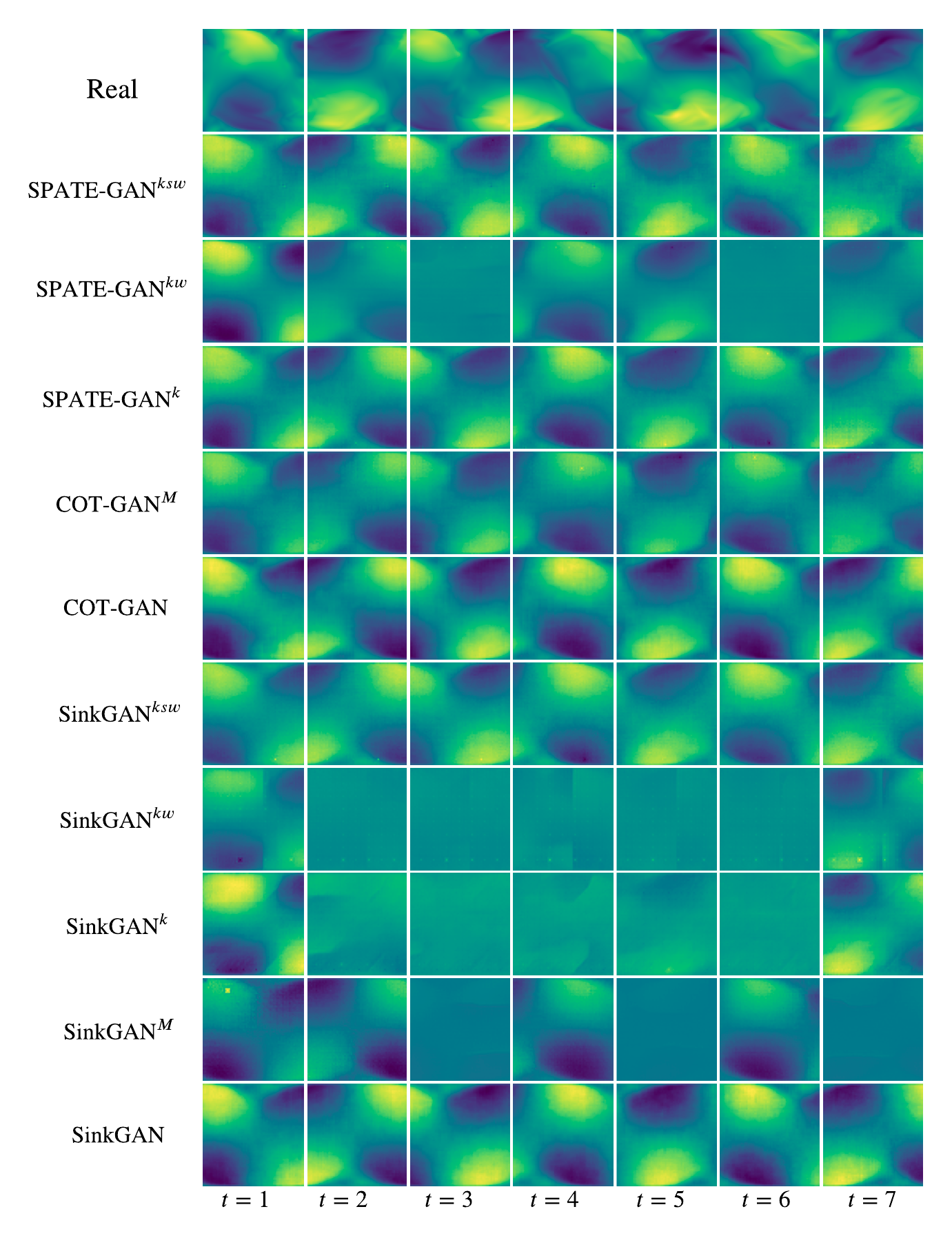}
    \caption{More selected samples for turbulent flow (TF) dataset.}
    \label{fig:tf_full}
\end{figure*} 